\documentclass[11pt,a4paper]{article}
\usepackage[hyperref]{acl2017}
\usepackage{times}
\usepackage{latexsym}
\usepackage{tabularx}
\usepackage{graphicx}
\usepackage{url}
\usepackage{xspace}
\usepackage{amsmath}
\usepackage{amssymb}
\usepackage{amsfonts}

\aclfinalcopy % Uncomment this line for the final submission
\def\aclpaperid{715} %  Enter the acl Paper ID here

\newcommand\squad{SQuAD\xspace}

\newcommand\lcurq{CuratedTREC\xspace}

\newcommand\wikim{WikiMovies\xspace}
\newcommand\wq{WebQuestions\xspace}

\newcommand\us{DrQA\xspace}
\newcommand\usr{Document Retriever\xspace}
\newcommand\usp{Document Reader\xspace}
\newcommand{\finalem}{70.0}
\newcommand{\finalf}{79.0}

\title{Reading Wikipedia to Answer Open-Domain Questions}
\author{Danqi Chen\thanks{\hspace{0.3em} Most of this work was done while DC was with Facebook AI Research.} \\
Computer Science\\ 
Stanford University\\
Stanford, CA 94305, USA\\
  {\tt danqi@cs.stanford.edu} \\\And
Adam Fisch, Jason Weston \& Antoine Bordes\\
Facebook AI Research\\
770 Broadway\\
New York, NY 10003, USA\\
{\tt \{afisch,jase,abordes\}@fb.com} \\}

\date{}

\begin{document}
\maketitle
\begin{abstract}
This paper proposes to tackle open-domain question answering using Wikipedia as the unique knowledge source: the answer to any factoid question is a text span in a Wikipedia article. This task of {\it machine reading at scale} combines the challenges of document retrieval (finding the relevant articles) with that of machine comprehension of text (identifying the answer spans from those articles). Our approach combines a search component based on bigram hashing and TF-IDF matching  with a multi-layer recurrent neural network model trained to detect answers in Wikipedia paragraphs. Our experiments on multiple existing QA datasets indicate that (1) both modules are highly competitive with respect to existing counterparts and (2) multitask learning using distant supervision on their combination is an effective complete system on this challenging task.
\end{abstract}

\section{Introduction}

This paper considers the problem of answering factoid questions in an open-domain setting using Wikipedia as the unique knowledge source, such as one does when looking for answers in an encyclopedia.
Wikipedia is a constantly evolving source of detailed information that could facilitate intelligent machines  --- if they are able to leverage its power.
Unlike knowledge bases (KBs) such as Freebase~\cite{bollacker2008freebase} or DBPedia~\cite{auer2007dbpedia}, which are easier for computers to process but too sparsely populated for open-domain question answering \cite{miller2016key}, Wikipedia contains up-to-date knowledge that humans are interested in. It is designed, however, for humans -- not machines -- to read.
%
%Our work treats Wikipedia as a collection of articles and does not rely on its internal graph structure. As a result, our approach is generic and could switch it for another collection of documents.
%
%Wikipedia is richer than a KB: no schema, great variability of answers, multiples languages.

Using Wikipedia articles as the knowledge source causes the task of question answering (QA) to combine the challenges of both large-scale open-domain QA and of machine comprehension of text. In order to answer any question, one must first retrieve the few relevant articles among more than 5 million items, and then scan them carefully to identify the answer.
We term this setting, {\it machine reading at scale} (MRS).
Our work treats Wikipedia as a collection of articles and does not rely on its internal graph structure. As a result, our approach is generic and could be switched to other collections of documents, books, or even daily updated newspapers.

Large-scale QA systems like IBM's DeepQA ~\cite{ferrucci2010building} rely on multiple sources to answer: besides Wikipedia, it is also paired with KBs, dictionaries, and even news articles, books, etc. As a result, such systems heavily rely on information redundancy among the sources to answer correctly. Having a single knowledge source forces the model to be very precise while searching for an answer as the evidence might appear only once. This challenge thus encourages research in the ability of a machine to read, a key motivation for the machine comprehension subfield and the creation of datasets such as SQuAD~\cite{rajpurkar2016squad}, CNN/Daily Mail~\cite{nips2015hermann} and CBT~\cite{hill2015goldilocks}.

However, those machine comprehension resources typically assume that a short piece of relevant text is already identified and given to the model, which is not realistic for building an open-domain QA system. In sharp contrast, methods that use KBs or information retrieval over documents have to employ search as an integral part of the solution. Instead MRS is focused on simultaneously maintaining the challenge of machine comprehension, which requires the deep understanding of text, while keeping the realistic constraint of searching over a large open resource.

\begin{figure*}
    \begin{center}
    \includegraphics[scale=0.42]{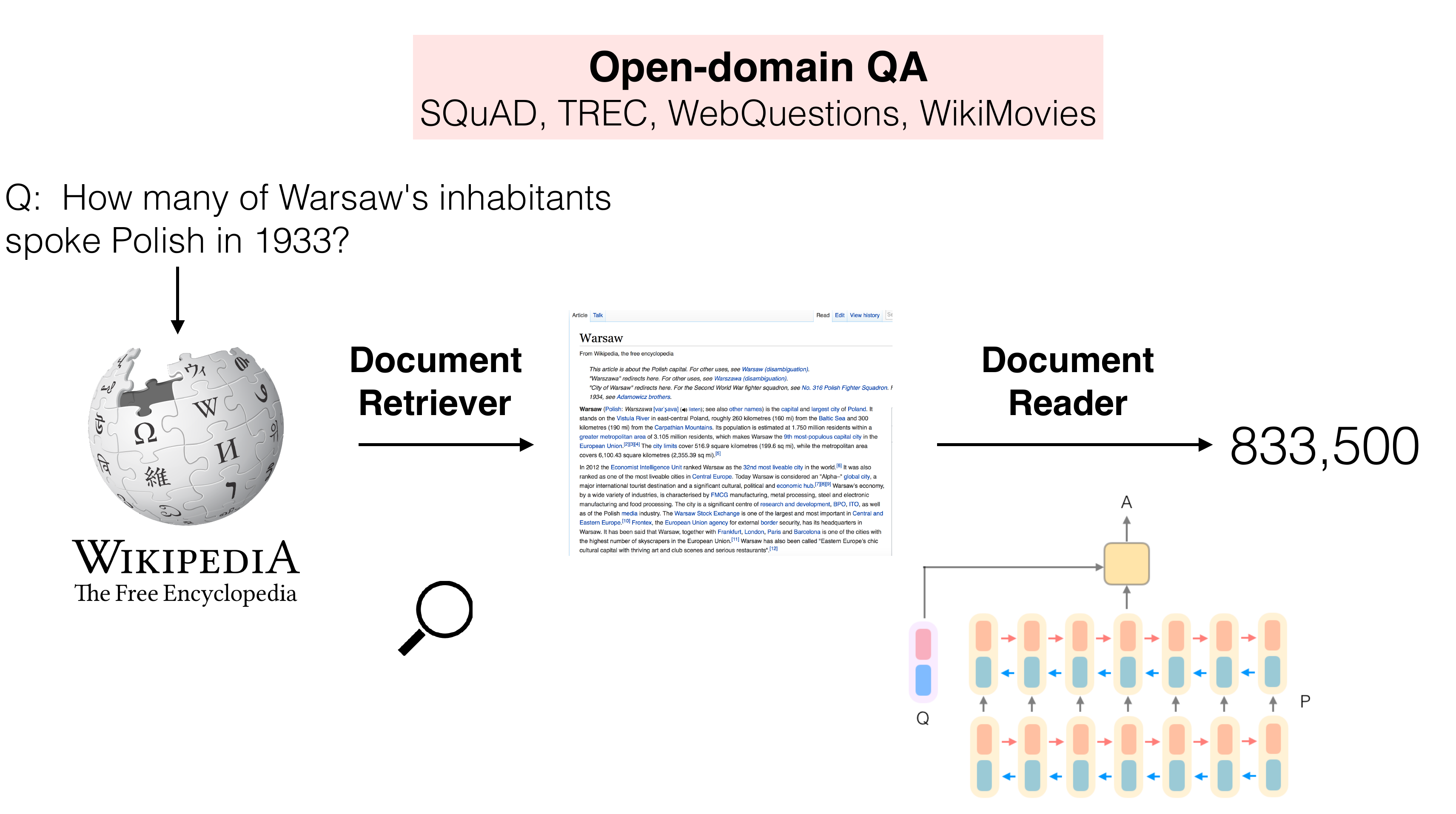}
    \end{center}
    \vspace{-1em}
    \caption{\label{fig:pip} An overview of our question answering system \us. }
\end{figure*}

%In that respect, QA from Wikipedia alone requires machines to have a deep %understanding of text as tested by recent resources like

%
%This also means that the answers are not restricted to be specific pre-defined entities, but can be any span of text, of any length and that one can not use the structure of KBs but must interpret plain text to discover responses.
%
%In that respect, QA from Wikipedia alone requires machines to have a deep understanding of text as tested by recent resources like SQuAD~\cite{rajpurkar2016squad}, QACNN~\cite{nips15_hermann} or CBT~\cite{hill2015goldilocks}.

In this paper, we show how multiple existing QA datasets can be used to evaluate MRS by requiring an open-domain system to perform well on all of them at once.
We develop \us, a strong system for question answering from Wikipedia composed of: (1)  \usr, a module using bigram hashing and TF-IDF matching  designed to, given a question, efficiently return a subset of relevant articles and (2) \usp, a multi-layer recurrent neural network machine comprehension
model trained to detect answer spans in those few returned  documents.
Figure~\ref{fig:pip} gives an illustration of \us.
%
%Section~\ref{sec:model} details \us.
%

Our experiments show that \usr outperforms the built-in Wikipedia search engine and that \usp reaches state-of-the-art results on the very competitive SQuAD benchmark \cite{rajpurkar2016squad}.
%
%Results across multiple datasets indicate that our full system, \us, is \textcolor{red}{\bf competitive} with YodaQA~\cite{baudivs2015yodaqa}, an open-source QA pipeline using multiple data sources and inspired by the DeepQA architecture. \textcolor{blue}{Mention combine different data sources for training?}
%
%We show that  combining multitask learning and distant supervision across multiple question answer datasets outperforms single task training when evaluating our full system across multiple benchmarks.
%
%
Finally, our full system is evaluated using multiple benchmarks. In particular, we show that performance is improved across all datasets through the use of multitask learning and distant supervision compared to single task training. %We arrive at a single system where the aim is to correctly respond to any question where Wikipedia can provide the answer.

% Our full system aims to correctly respond to any question where Wikipedia can provide the answer, as such it is evaluated using multiple benchmarks. We find that performance is improved across almost all datasets through the use of multitask learning and distant supervision compared to single task training. We arrive at a single system where the aim is to correctly respond to any question where Wikipedia can provide the answer.

%The paper is organized as follows. Section~\ref{sec:rwork} discusses related work  and Section~\ref{sec:model} introduces our system, \us. Sections~\ref{sec:data} describes the various data sources used in this paper. Finally, Section~\ref{sec:exp} presents our experiments on SQuAD, for the components Document Retriever and Document Reader, and for the full system in the open-domain QA setting.

\section{Related Work} \label{sec:rwork}

Open-domain QA was originally defined as finding answers in collections of unstructured documents, following the setting of the annual TREC competitions\footnote{\url{http://trec.nist.gov/data/qamain.html}}.
With the development of KBs, many recent innovations have occurred in the context of QA from KBs with the creation of resources like WebQuestions \cite{berant2013semantic} and SimpleQuestions \cite{bordes2015large} based on the Freebase KB \cite{bollacker2008freebase},
or on automatically extracted KBs, e.g., OpenIE triples and NELL \cite{fader2014open}.
However, KBs have inherent limitations (incompleteness, fixed schemas) that
motivated researchers to return to the original setting of answering from raw text.

A second motivation to cast a fresh look at this problem is that of machine comprehension of text, i.e., answering questions after reading a short text or story. That subfield has made considerable progress recently
thanks to new deep learning architectures like attention-based and memory-augmented neural networks \cite{bahdanau2015neural,weston2015memory,graves2014neural} and release of new training and evaluation datasets like QuizBowl \cite{iyyer2014neural}, CNN/Daily Mail based on news articles \cite{nips2015hermann}, CBT based on children books \cite{hill2015goldilocks}, or SQuAD \cite{rajpurkar2016squad} and WikiReading \cite{hewlett2016wiki}, both based on Wikipedia.
An objective of this paper is to test how such new methods can perform in an open-domain QA framework.

QA using Wikipedia as a resource has been explored previously.
\citet{ryu2014open} perform open-domain QA using a Wikipedia-based knowledge model. They combine article content with multiple other answer matching modules based on different types of semi-structured knowledge such as infoboxes, article structure, category structure, and definitions.
 Similarly, \citet{Ahn04} also combine Wikipedia as a text resource with other resources, in this case with information retrieval over other documents.
\citet{buscaldi2006mining} also mine knowledge from Wikipedia for QA. Instead of using it as a resource for seeking answers to questions, they focus on validating answers returned by their QA system, and use Wikipedia categories for determining a set of patterns that should fit with the expected answer.
In our work, we consider the comprehension of text only, and use Wikipedia text documents as the sole resource in order to emphasize the task of machine reading at scale, as described in the introduction.

%"Mining knowledge from wikipedia for the question answering task" \cite{buscaldi2006mining}
%"Previous works considered Wikipedia as a resource where to look for the answers to the questions. We focused on some different aspects of the problem, such as the validation of the answers as returned by our Question Answering System and on the use of Wikipedia “categories” in order to determine a set of patterns that should fit with the expected answer."

%"Open domain question answering using Wikipedia-based knowledge model" \cite{ryu2014open}
% "We suggest multiple answer matching modules based on different types of semi-structured knowledge sources of Wikipedia, including article content, infoboxes, article structure, category structure, and definitions."

There are a number of highly developed full pipeline QA approaches using either the Web, as does QuASE \cite{sun2015open}, or Wikipedia as a resource, as do Microsoft's AskMSR \cite{brill2002askmsr},
IBM's DeepQA \cite{ferrucci2010building} and YodaQA \cite{baudivs2015yodaqa,baudivs2015modeling} --- the latter of which is open source and hence reproducible for comparison purposes.
AskMSR is a search-engine based QA system that relies on ``data redundancy rather than sophisticated linguistic analyses of either questions or candidate answers'', i.e., it does not focus on machine comprehension, as we do. DeepQA is a very sophisticated system that relies on both unstructured information including text documents as well as structured data such as KBs, databases and ontologies to generate candidate answers or vote over evidence. YodaQA is an open source system modeled after DeepQA, similarly combining websites, information extraction, databases and Wikipedia in particular. Our comprehension task is made more challenging by only using a single resource. Comparing against these methods provides a useful datapoint for an ``upper bound'' benchmark on performance.

%Full pipeline
%ASkMSR \cite{brill2002askmsr}
% "The system differs from most question answering systems in its dependency on data redundancy rather than sophisticated linguistic analyses of either questions or candidate answers. "

%DeepQA \cite{ferrucci2010building}
% "the majority of evidence analysis in DeepQA is focused on unstructured information (e.g., natural-language documents), several components in the DeepQA system use structured data (e.g., databases, knowledge bases, and ontologies) to generate potential candidate answers or find additional evidence"
% Interesting Paper list: http://ieeexplore.ieee.org/xpl/tocresult.jsp?isnumber=6177717&cm_mc_uid=64431735148814861341005&cm_mc_sid_50200000=148615753

%YodaQA \cite{baudivs2015yodaqa,baudivs2015modeling}
% Wikipedia, websites, NLP, information extraction problem, Databases (RDF graph): current work, today’s
% topic
% DBpedia (Wikipedia infoboxes)
% Freebase (Google Knowledge Graph)

%latest TREC'16 \cite{mackinnon2007complex}

Multitask learning \cite{caruana1998multitask} and task transfer have a rich history in machine learning (e.g., using ImageNet in the computer vision community \cite{huh2016makes}),
as well as in NLP in particular \cite{Collobert08}.
Several works have attempted to combine multiple QA training datasets via multitask learning to (i) achieve improvement across the datasets via task transfer; and (ii) provide a single general system capable of asking different kinds of questions due to the inevitably different data distributions across the source datasets. \citet{fader2014open} used WebQuestions, TREC and WikiAnswers with four KBs as knowledge sources and reported improvement on the latter two datasets through multitask learning.
\citet{bordes2015large} combined WebQuestions and SimpleQuestions using distant supervision with Freebase as the KB to give slight improvements on both datasets, although poor performance was reported when training on only one dataset and testing on the other, showing that task transfer is indeed a challenging subject; see also \cite{kadlecparticular} for a similar conclusion.
Our work follows similar themes, but in the setting of having to retrieve and then read text documents, rather than using a KB, with positive results.
%, rather than retrieving and reading text as we do here,

% fader2014open
% WebQuestions, TREC and WikiAnswers

\section{Our System: \us} \label{sec:model}

In the following we describe our system \us for MRS which consists of two components: (1) the Document Retriever module for finding relevant articles and (2) a machine comprehension model, \usp, for extracting answers from a single document or a small collection of documents.
% A model made of two parts following a standard open QA pipeline\\
% - tfidf\\
% - \usp\\
% - simple aggregation.
\subsection{\usr} \label{sec:irmodel}
Following classical QA systems, we use an efficient (non-machine learning) document retrieval system to first narrow our search space and focus on reading only articles that are likely to be relevant. A simple inverted index lookup followed by term vector model scoring performs quite well on this task for many question types, compared to the built-in ElasticSearch based Wikipedia Search API \cite{gormley2015elasticsearch}. Articles and questions are compared as TF-IDF weighted bag-of-word vectors.
%
%weighted with the probabilistic TF-IDF:
%\begin{equation*}
%t_{f}\log{\frac{N-n_t+0.5}{n_t+0.5}}  % this equation is probably trivial
%\end{equation*}
%with $t_f$
%
We further improve our system by taking local word order into account with n-gram features. Our best performing system uses bigram counts while preserving speed and memory efficiency by using the hashing of \cite{weinberger2009feature} to map the bigrams to $2^{24}$ bins with an unsigned \emph{murmur3} hash.

We use \usr as the first part of our full model, by setting it to return 5 Wikipedia articles given any question. Those articles are then processed by \usp.

\subsection{Document Reader}
Our \usp model is inspired by the recent success of neural network models on machine comprehension tasks, in a similar spirit to the \emph{AttentiveReader} described in \cite{nips2015hermann,chen2016thorough}.

%\begin{figure}
%    \begin{center}
%    \includegraphics[scale=0.3]{figures/nn.pdf}
%  \end{center}
%  \caption{An illustration of \usp.}
%\end{figure}

Given a question $q$ consisting of $l$ tokens $\{q_1, \ldots, q_l\}$ and a
document or a small set of documents of $n$ paragraphs where a single paragraph $p$ consists of $m$ tokens $\{p_1, \ldots, p_m\}$,  we develop an RNN model that we apply to each paragraph in turn and then finally aggregate the predicted answers. Our method works as follows:

\paragraph{\textbf{Paragraph encoding}} We first represent all tokens $p_i$ in a paragraph $p$
as a sequence of feature vectors $\tilde{\mathbf{p}}_i \in \mathbb{R}^d$ and pass them as the input to a recurrent neural network and thus obtain:
\begin{equation*}
    \{\mathbf{p}_1, \ldots, \mathbf{p}_m\} = \text{RNN}(\{\tilde{\mathbf{p}}_1, \ldots, \tilde{\mathbf{p}}_m \}),
\end{equation*}
where $\mathbf{p}_i$ is expected to encode useful context information around token $p_i$. Specifically, we choose to use a multi-layer bidirectional long short-term memory network (LSTM), and take $\mathbf{p}_i$ as the concatenation of each layer's hidden units in the end.

The feature vector $\tilde{\mathbf{p}}_i$ is comprised of the following parts:
\begin{itemize}
    \item
%    \vspace{-0.5em}
    \emph{Word embeddings}: $f_{emb}(p_i) = \mathbf{E}(p_i)$. We use the 300-dimensional Glove word embeddings trained from 840B Web crawl data \cite{pennington2014glove}. We keep most of the pre-trained word embeddings fixed and only fine-tune the $1000$ most frequent question words because the representations of some key words such as \textit{what}, \textit{how}, \textit{which}, \textit{many} could be crucial for QA systems.
    \item
%    \vspace{-0.5em}
    \emph{Exact match}: $f_{exact\_match}(p_i) = \mathbb{I}(p_i \in q)$. We use three simple binary features, indicating whether $p_i$ can be exactly matched to one question word in $q$, either in its original, lowercase or lemma form. These simple features turn out to be extremely helpful, as we will show in Section~\ref{sec:exp}.
    \item
%      \vspace{-0.5em}
    \emph{Token features}: \\ $f_{token}(p_i) = (\text{POS}(p_i), \text{NER}(p_i), \text{TF}(p_i))$. We also add a few manual features which reflect some properties of token $p_i$ in its context, which include its part-of-speech (POS) and named entity recognition (NER) tags and its (normalized) term frequency (TF).
    \item
%    \vspace{-0.5em}
    \emph{Aligned question embedding}: \\ Following \cite{lee2016learning} and other recent works, the last part we incorporate is an aligned question embedding $f_{align}(p_i) = \sum_j{a_{i, j} \mathbf{E}(q_j)}$, where the attention score $a_{i, j}$ captures the similarity between $p_i$ and each question words $q_j$. Specifically, $a_{i, j}$ is computed by the dot products between nonlinear mappings of word embeddings: 
    \begin{equation*}
    	a_{i, j} = \frac{\exp\left(\alpha(\mathbf{E}(p_i)) \cdot \alpha(\mathbf{E}(q_{j}))\right)}{\sum_{j'}{\exp\left(\alpha(\mathbf{E}(p_i)) \cdot \alpha(\mathbf{E}(q_{j'}))\right)}},
    \end{equation*} and $\alpha(\cdot)$ is a single dense layer with ReLU nonlinearity. Compared to the \emph{exact match} features, these features add soft alignments between similar but non-identical words (e.g., \textit{car} and \textit{vehicle}).
\end{itemize}

% \textcolor{red}{I would pose this as searching for paraphrasing, with an emphasis on how simple word match features are the main driver of performance. Then as an added bonus -- to get an extra .8\% -- we can add the soft question word attention, such as the mechanism of lee et al. IMO this switches the focus away from the question embeddings, like you wanted.}

\paragraph{\textbf{Question encoding}} The question encoding is simpler, as we only apply another recurrent neural network on top of the word embeddings of $q_i$ and combine the resulting hidden units into one single vector: $\{\mathbf{q}_1, \ldots, \mathbf{q}_l\} \rightarrow \mathbf{q}$. We compute $\mathbf{q} = \sum_j{b_j \mathbf{q}_j}$ where $b_j$ encodes the importance of each question word:
\begin{equation*}
	b_j = \frac{\exp(\mathbf{w} \cdot \mathbf{q}_j)}{\sum_{j'}{\exp(\mathbf{w} \cdot \mathbf{q}_{j'})}},
\end{equation*}
and $\mathbf{w}$ is a weight vector to learn.

\paragraph{\textbf{Prediction}} At the paragraph level, the goal is to predict the span of tokens that is most likely the correct answer. We take the
the paragraph vectors $\{\mathbf{p}_1, \ldots, \mathbf{p}_m\}$ and the question vector $\mathbf{q}$ as input, and simply train two classifiers independently for predicting the two ends of the span. Concretely, we use a bilinear term to capture the similarity between $\mathbf{p}_i$ and $\mathbf{q}$ and compute the probabilities of each token being start and end as:
\begin{eqnarray*}
P_{start}(i) & \propto & \exp \left(\mathbf{p}_i \mathbf{W}_{s} \mathbf{q}\right) \\
P_{end}(i) & \propto & \exp \left(\mathbf{p}_i \mathbf{W}_{e} \mathbf{q}\right)
\end{eqnarray*}
During prediction, we choose the best span from token $i$ to token $i'$ such that $i \leq i' \leq i + 15$ and $P_{start}(i) \times P_{end}(i')$ is maximized.
To make scores compatible across paragraphs in one or several retrieved documents, we use the unnormalized exponential and take argmax over all considered paragraph spans for our final prediction.

\begin{table*}
\begin{center}
\small
\begin{tabularx}{\textwidth}{l|p{5.2cm}|p{7.8cm}}
\hline
\bf Dataset & \bf Example & \bf Article / Paragraph \\
\hline
\squad & {\bf Q}: How many provinces did the Ottoman empire contain in the 17th century? \newline  {\bf A}: 32 &  {\bf Article}: Ottoman Empire \newline {\bf Paragraph}: ... At the beginning of the 17th century the empire contained \textcolor{blue}{32} provinces and numerous vassal states. Some of these were later absorbed into the Ottoman Empire, while others were granted various types of autonomy during the course of centuries.\\
\hline
%\curq & \textcolor{red}{TODO: I don't have this dataset.. }\\
%\hline
\lcurq & {\bf Q}: What U.S. state's motto is ``Live free or Die''? \newline {\bf A}: New Hampshire & {\bf Article}: Live Free or Die \newline {\bf Paragraph}: "Live Free or Die" is the official motto of the U.S. state of \textcolor{blue}{New Hampshire}, adopted by the state in 1945. It is possibly the best-known of all state mottos, partly because it conveys an assertive independence historically found in American political philosophy and partly because of its contrast to the milder sentiments found in other state mottos.\\
\hline
\wq  & {\bf Q}: What part of the atom did Chadwick discover?$^\dagger$  \newline {\bf A}: neutron  & {\bf Article}: Atom \newline {\bf Paragraph}: ... The atomic mass of these isotopes varied by integer amounts, called the whole number rule. The explanation for these different isotopes awaited the discovery of the \textcolor{blue}{neutron}, an uncharged particle with a mass similar to the proton, by the physicist James Chadwick in 1932.  ... \\
\hline
\wikim & {\bf Q}: Who wrote the film Gigli? \newline {\bf A}: Martin Brest &  {\bf Article}: Gigli \newline {\bf Paragraph}: Gigli is a 2003 American romantic comedy film written and directed by \textcolor{blue}{Martin Brest} and starring Ben Affleck, Jennifer Lopez, Justin Bartha, Al Pacino, Christopher Walken, and Lainie Kazan. \\
\hline
\end{tabularx}
\end{center}
\caption{\label{tab:ex}Example training data from each QA dataset. In each case we show an associated paragraph where distant supervision (DS) correctly identified the answer within it, which is highlighted.} % $^\dagger$: The original questions in WebQuestions are all lowercased.}
\end{table*}

\section{Data} \label{sec:data}

Our work relies on three types of data: (1) Wikipedia that serves as our knowledge source for finding answers, (2) the SQuAD dataset which is our main resource to train \usp and (3) three more QA datasets (CuratedTREC, WebQuestions and WikiMovies)  that in addition to SQuAD, are used to test the open-domain QA abilities of our full system, and to evaluate the ability of our model to learn from multitask learning and distant supervision.
Statistics of the datasets are given in Table~\ref{tab:data-stats}.

\begin{table}[h]
\begin{center}
\begin{tabular}{l|c@{\,\,}c@{\,\,}c}
\hline
\bf Dataset & \multicolumn{2}{c}{\bf Train} & \bf Test  \\
& Plain & DS &  \\
\hline
\squad &  87,599 & 71,231 & 10,570$^{\dagger}$ \\
%\curq & 430$^{*}$ & - & 430 \\
\lcurq &  1,486$^{*}$ & 3,464 & 694 \\
\wq &  3,778$^{*}$ & 4,602 & 2,032 \\
\wikim &  96,185$^{*}$ & 36,301 & 9,952 \\
\hline
\end{tabular}
\end{center}
\caption{\label{tab:data-stats} Number of questions for each dataset used in this paper. DS: distantly supervised training data.
$^{*}$: These training sets are not used as is because no paragraph is associated with each question.
$^{\dagger}$: Corresponds to SQuAD development set.}
\end{table}

\subsection{Wikipedia (Knowledge Source)}
We use the 2016-12-21 dump\footnote{\url{https://dumps.wikimedia.org/enwiki/latest}} of English Wikipedia for all of our full-scale experiments as the knowledge source used to answer questions. For each page, only the plain text is extracted and all structured data sections such as lists and figures are stripped.\footnote{We use the WikiExtractor script: \url{https://github.com/attardi/wikiextractor}.} After discarding internal disambiguation, list, index, and outline pages, we retain 5,075,182 articles consisting of 9,008,962 unique uncased token types.

\subsection{SQuAD}

The Stanford Question Answering Dataset (SQuAD) \cite{rajpurkar2016squad} is a dataset for machine comprehension based on Wikipedia.
The dataset contains 87k examples for training and 10k for development, with a large hidden test set which can only be accessed by the SQuAD creators.
Each example is composed of a paragraph extracted from a Wikipedia article and an associated human-generated question. The answer is always a span from this paragraph and a model is given credit if its predicted answer matches it. Two evaluation metrics are used: exact string match (EM) and F1 score, which measures the weighted average of precision and recall at the token level.

In the following, we use SQuAD for training and evaluating our Document Reader for the standard machine comprehension task given the relevant paragraph as defined in \cite{rajpurkar2016squad}.
For the task of evaluating open-domain question answering over Wikipedia, we use the SQuAD development set QA pairs only, and we ask systems to uncover the correct answer spans \emph{without} having access to the associated paragraphs. That is, a model is required to answer a question given the whole of Wikipedia as a resource; it is {\em not} given the relevant paragraph as in the standard SQuAD setting.

\subsection{Open-domain QA Evaluation Resources}\label{sec:othersources}

SQuAD is one of the largest general purpose QA datasets currently available.
SQuAD questions have been collected via a process involving showing a paragraph to each human annotator and asking them to write a question. As a result, their distribution is quite specific.
We hence propose to train and evaluate our system on other datasets developed for open-domain QA that have been constructed in different ways (not necessarily in the context of answering from Wikipedia).

\paragraph{\lcurq} This dataset is based on the
benchmarks from the TREC QA tasks that have been curated by \citet{baudivs2015modeling}.
%
%It contains 430 training and 430 testing questions.
%
We use the large version, which contains a total of 2,180 questions extracted from the datasets from TREC 1999, 2000, 2001 and 2002.\footnote{This dataset is available at \url{https://github.com/brmson/dataset-factoid-curated}.}

\paragraph{\wq} Introduced in \cite{berant2013semantic}, this dataset is built to answer questions from the Freebase KB. It was created by crawling questions through the Google Suggest API, and then obtaining answers using Amazon Mechanical Turk.
We convert each answer to text by using entity names so that the dataset does not reference Freebase IDs and is purely made of plain text question-answer pairs.

\paragraph{\wikim} This dataset, introduced in \cite{miller2016key}, contains 96k question-answer pairs in the domain of movies. Originally created from the OMDb and MovieLens databases, the examples are built such that they can also be answered by using a subset of Wikipedia as the knowledge source (the title and the first section of articles from the movie domain).

\begin{table}[t]
\begin{center}
\normalsize
\begin{tabular}{l|c|cc}
\hline
\bf Dataset &  \bf Wiki & \multicolumn{2}{c}{\bf Doc. Retriever} \\
&  \bf Search  & plain &  +bigrams  \\
\hline
\squad & 62.7 &  76.1 & \bf 77.8 \\
%\curq  & 82.8 & 84.2 & \bf 85.6 \\
\lcurq & 81.0 & 85.2 & \bf 86.0 \\
\wq &    73.7 & \bf 75.5 & 74.4 \\
\wikim & 61.7 &  54.4 &  \bf 70.3 \\
\hline
\end{tabular}
\end{center}
\caption{\label{tab:ir-res} Document retrieval results. \% of questions for which the answer segment appears in one of the top 5 pages returned by the method. }
\end{table}

\subsection{Distantly Supervised Data} \label{sec:ds}

All the QA datasets presented above contain training portions, but CuratedTREC, WebQuestions
and WikiMovies only contain question-answer pairs, and not
an associated document or paragraph as in SQuAD, and hence
cannot be used for training \usp directly.
%
%which cannot be used for training \usp directly because not %paired with associated paragraphs, as for SQuAD.
%
Following previous work on distant supervision (DS) for relation extraction \cite{mintz2009distant}, we use a procedure to automatically associate paragraphs to such training examples, and then add these examples to our
training set.
%in the hope that these additional training examples might improve \us, especially evaluating on other resources besides SQuAD.

We use the following process for each question-answer pair to build our training set.
First, we run \usr on the question to retrieve the top 5 Wikipedia articles.
All paragraphs from those articles without an exact match of the known answer are directly discarded.
All paragraphs  shorter than 25 or longer than 1500  characters are also filtered out.
If any named entities are detected in the question, we remove any paragraph that does not contain them at all.
For every remaining paragraph in each retrieved page, we score all positions that match an answer using unigram and bigram overlap between the question and a 20 token window, keeping up to the top 5 paragraphs with the highest overlaps. If there is no paragraph with non-zero overlap, the example is discarded; otherwise we add each found pair to our DS training dataset. Some examples are shown in Table~\ref{tab:ex} and data statistics are given in Table~\ref{tab:data-stats}.

Note that we can also generate additional DS data for SQuAD by trying to find mentions of the answers not just in the paragraph provided, but also from other pages or the same page that the given paragraph was in. We observe that around half of the DS examples come from pages outside of the articles used in SQuAD.

\begin{table*}[t]
\begin{center}
\begin{tabular}{l|c@{\,\,\,\,}cc@{\,\,\,\,}c}
\hline
 \bf Method &  \multicolumn{2}{c}{\bf Dev} & \multicolumn{2}{c}{\bf Test} \\
&  EM & F1 & EM & F1 \\
\hline
%\bf Ensembles & & & & \\
%Multi-Perspect. Match. &  73.3 & 81.1 & 73.8 & 81.3 \\
%R-Net &   - & - & 75.9 & 82.9 \\
%\usp & -  & - & - & - \\
%\hline
%\hline
%\bf Single models & & & & \\
Dynamic Coattention Networks \cite{xiong2016dynamic} & 65.4 & 75.6 & 66.2 & 75.9 \\
Multi-Perspective Matching \cite{wang2016multi}$^\dagger$   & 66.1 & 75.8 & 65.5 & 75.1 \\
BiDAF \cite{seo2016bidirectional} & 67.7 & 77.3 & 68.0 & 77.3 \\
R-net$^\dagger$   & n/a & n/a & 71.3 & 79.7 \\
\hline
\us (Our model, Document Reader Only) & \bf 69.5 & \bf 78.8 & {\finalem} & {\finalf} \\
% \multicolumn{1}{r|}{\it - with TrainAuto data} & -  & - & - & -  \\
% \multicolumn{1}{r|}{\it - without features} & -  & - & - & -   \\
\hline
\end{tabular}
\end{center}
\caption{\label{tab:squad-res} Evaluation results on the SQuAD dataset (single model only). $^\dagger$: Test results reflect the SQuAD leaderboard {\small(\url{https://stanford-qa.com})} as of Feb 6, 2017.}
\end{table*}

\section{Experiments} \label{sec:exp}

This section first presents evaluations of our Document Retriever and Document Reader modules separately, and then describes tests of their combination, \us, for open-domain QA on the full Wikipedia.

\subsection{Finding Relevant Articles}
We first examine the performance of our \usr module on all the QA datasets. Table~\ref{tab:ir-res} compares the performance of the two approaches described in Section~\ref{sec:irmodel} with that of the Wikipedia Search Engine\footnote{We use the Wikipedia Search API \url{https://www.mediawiki.org/wiki/API:Search}.} for the task of finding articles that contain the answer given a question.
Specifically, we compute the ratio of questions for which the text span of any of their associated answers appear in at least one the top 5 relevant pages returned by each system.
Results on all datasets indicate that our simple approach outperforms Wikipedia Search, especially with bigram hashing.
We also compare doing retrieval with Okapi BM25 or by using cosine distance in the word embeddings space (by encoding questions and articles as bag-of-embeddings), both of which we find performed worse.

\subsection{Reader Evaluation on SQuAD}
Next we evaluate our \usp component on the standard SQuAD evaluation \cite{rajpurkar2016squad}.
%In the subsequent subsection, we will explore how to combine more training resources and evaluate on all the QA datasets.
\paragraph{Implementation details}
We use $3$-layer bidirectional LSTMs with $h = 128$ hidden units for both paragraph and question encoding.
We apply the Stanford CoreNLP toolkit \cite{manning2014stanford} for tokenization and also generating lemma, part-of-speech, and named entity tags.

Lastly, all the training examples are sorted by the length of paragraph and divided into mini-batches of $32$ examples each. We use \emph{Adamax} for optimization as described in \cite{kingma2014adam}. Dropout with $p = 0.3$ is applied to word embeddings and all the hidden units of LSTMs.
\paragraph{Result and analysis} Table \ref{tab:squad-res} presents our evaluation results on both development and test sets. SQuAD has been a very competitive machine comprehension benchmark since its creation and we only list the best-performing systems in the table. Our system (single model) can achieve {\finalem}\% exact match and {\finalf}\% F1 scores on the test set, which surpasses all the published results and can match the top performance on the SQuAD leaderboard at the time of writing. Additionally, we think that our model is conceptually simpler than most of the existing systems. We conducted an ablation analysis on the feature vector of paragraph tokens. As shown in Table \ref{tab:feature-ablation} all the features contribute to the performance of our final system. Without the aligned question embedding feature (only word embedding and a few manual features), our system is still able to achieve F1 over $77\%$. More interestingly, if we remove both $f_{aligned}$ and $f_{exact\_match}$, the performance drops dramatically, so we conclude that both features play a similar but complementary role in the feature representation related to the paraphrased nature of a question vs. the context around an answer.

\begin{table}
	\begin{center}
	\begin{tabular}{l | l}
    \hline
    \bf Features & \bf F1\\
    \hline
    Full & 78.8 \\
    \hline
    No $f_{token}$ & 78.0 (-0.8)\\
    No $f_{exact\_match}$ & 77.3 (-1.5)\\
    No $f_{aligned}$ & 77.3 (-1.5)\\
    No $f_{aligned}$ and $f_{exact\_match}$ & 59.4 (-19.4) \\
    \hline
    \end{tabular}
    \end{center}
    \caption{\label{tab:feature-ablation}Feature ablation analysis of the paragraph representations of our Document Reader. Results are reported on the SQuAD development set.}
\end{table}

\begin{table*}[ht]
\begin{center}
\begin{tabular}{l|c|ccc}
\hline
\bf Dataset &  \bf YodaQA &  \multicolumn{3}{c}{\bf \us} \\
&   &  SQuAD &  +Fine-tune (DS) & +Multitask (DS) \\
\hline
\squad~{\small\it(All Wikipedia)}&  n/a & 27.1 & 28.4 & 29.8\\
% \curq & 36.5 & 21.2 & 24.6   \\
\lcurq & 31.3 & 19.7 & 25.7 & 25.4   \\
\wq & 39.8 & 11.8 & 19.5 & 20.7  \\
\wikim & n/a & 24.5 & 34.3 & 36.5  \\
\hline
%Micro-average & n/a & 24.6* & \bf 27.9* \\
%Macro-average & n/a &  22.4* & \bf 25.4* \\
%\hline
\end{tabular}
\end{center}
\caption{\label{tab:full-pip-res} Full Wikipedia results. Top-1 exact-match accuracy (in \%, using SQuAD eval script). +Fine-tune (DS): \usp models trained on SQuAD and fine-tuned on each DS training set independently. +Multitask (DS): \usp single model trained on SQuAD and all the distant supervision (DS) training sets jointly. YodaQA results are extracted from {\small \url{https://github.com/brmson/yodaqa/wiki/Benchmarks}} and use additional resources such as Freebase and
DBpedia, see Section \ref{sec:rwork}.}
\end{table*}

\subsection{Full Wikipedia Question Answering}

Finally, we assess the performance of our full system \us for answering open-domain questions using the four datasets introduced in Section~\ref{sec:data}.
%
%\paragraph{Settings}
We compare three versions of \us which evaluate the impact of using distant supervision and multitask learning across the training sources provided to \usp (\usr remains the same for each case):
%differ by the training sources that \usp is provided with
\begin{itemize}
\item SQuAD:
A single \usp model is trained on the SQuAD training set only and used on all evaluation sets.
\vspace{-1mm}
\item
Fine-tune (DS): A \usp model is pre-trained on SQuAD and then fine-tuned for each dataset independently using its distant supervision (DS) training set.
\vspace{-1mm}
\item
Multitask (DS): A single \usp model is jointly trained on the SQuAD training set and {\em all} the DS sources.
\end{itemize}

For the full Wikipedia setting we use a streamlined model that does not use the CoreNLP parsed $f_{token}$ features or lemmas for $f_{exact\_match}$. We find that while these help for more exact paragraph reading in SQuAD, they don't improve results in the full setting. Additionally, WebQuestions and WikiMovies provide a list of candidate answers (e.g., 1.6 million Freebase entity strings for WebQuestions) and we restrict the answer span must be in this list during prediction.

\paragraph{Results}
Table~\ref{tab:full-pip-res} presents the results.
Despite the difficulty of the task compared to machine comprehension (where you are given the right paragraph) and unconstrained QA (using redundant resources), \us still provides reasonable performance across all four datasets.

We are interested in a single, full system that can answer any question using Wikipedia. The single model trained only on SQuAD is outperformed on all four of the datasets by the multitask model that uses distant supervision.
However performance when training on SQuAD alone is not far behind, indicating that task transfer is occurring.
%It is worth noting that adding automatically generated DS data helps on SQuAD itself, even though we already have a large amount of labeled training data. Unfortunately this does not translate into better performance in the comprehension setting of Table~\ref{tab:squad-res}.
%The majority of the improvement from SQuAD to Multitask (DS), however, is from the introduction of extra in-domain data. We see this in the improvements that fine-tuning on each dataset alone using DS also gives. Still, the Multitask (DS) system is the best single model for our overall goal.
The majority of the improvement from SQuAD to Multitask (DS) however is likely not from task transfer as fine-tuning on each dataset alone using DS also gives improvements, showing that is is the introduction of extra data in the same domain that helps. Nevertheless, the best single model that we can find is our overall goal, and that is the Multitask (DS) system.

%except for \wq that is perhaps too Freebase-specific.
%
%It is worth noting that adding automatically generated DS data can help on SQuAD itself, where we already have a large amount of labeled training data. This is only the case in the fine-tuning case, and unfortunately does not translate into better performance in the vanilla test setting of Table~\ref{tab:squad-res}.

We compare to an unconstrained QA system using redundant resources (not just Wikipedia), YodaQA \cite{baudivs2015yodaqa}, giving results which were previously reported on \lcurq and \wq. Despite the increased difficulty of our task, it is reassuring that our performance is not too far behind on \lcurq (31.3 vs. 25.4). The gap is slightly bigger on \wq, likely because this dataset was created from the specific structure of Freebase which YodaQA uses directly.

\us's performance on SQuAD compared to its Document Reader component on machine comprehension in Table \ref{tab:squad-res} shows a large drop (from 69.5 to 27.1)
%due to the paragraph no longer being provided as part of the task.
as we now are given Wikipedia to read, not a single paragraph.
Given the correct document (but not the paragraph) we can achieve 49.4, indicating many false positives
come from highly topical sentences.
This is despite the fact that the Document Retriever works relatively well (77.8\% of the time retrieving the answer, see Table \ref{tab:ir-res}).
It is worth noting that a large part of the drop comes from the nature of the SQuAD questions. They were written with a specific paragraph in mind, thus their language can be ambiguous when the context is removed. Additional resources other than SQuAD, specifically designed for MRS, might be needed to go further.

\section{Conclusion} \label{sec:conc}

We studied the task of machine reading at scale, by using Wikipedia as the unique knowledge source for open-domain QA. Our results indicate that MRS is a key challenging task for researchers to focus on. Machine comprehension systems alone cannot solve the overall task. Our method integrates search,
distant supervision, and multitask learning to provide an effective complete system.
Evaluating the individual components as well as the full system across multiple benchmarks showed the efficacy of our approach.

Future work should aim to improve over our \us system.
Two obvious angles of attack are:  (i) incorporate the fact that Document Reader aggregates over multiple paragraphs and documents directly in the training, as it currently trains on paragraphs independently; and (ii) perform end-to-end training across the Document Retriever and Document Reader pipeline, rather than independent systems.
%; including trainable parameters in the Retriever module could add more flexibility and power.

\subsection*{Acknowledgments}
The authors thank Pranav Rajpurkar for testing Document Reader on the test set of SQuAD.

%\bibliography{acl2017}
\bibliography{fullwiki}
\bibliographystyle{acl_natbib}

\end{document}